\documentclass[11pt]{article}

\usepackage{amsmath}
\usepackage{amsthm}
\usepackage{amssymb}
\usepackage{mathptmx}      

\usepackage{pdfpages}

\usepackage{fancyhdr}
\fancyheadoffset[LE,RO]{\marginparsep+\marginparwidth}
\newcounter{remno} 
{}

\newfont{\mfoo}{cmssdc10 scaled\magstep1}
\newfont{\mfo}{cmtt9 scaled\magstep1}

\usepackage[numbers]{natbib}

\newcommand{\argmax}{\mbox{\,\rm arg\,max}}

\newcommand*{\vbig}[1]{\vcenter{\hbox{\scalebox{1.2}{\ensuremath#1}}}}
\newcommand*{\vbigg}[1]{\vcenter{\hbox{\scalebox{1.3}{\ensuremath#1}}}}
\newcommand*{\vbiggg}[1]{\vcenter{\hbox{\scalebox{1.5}{\ensuremath#1}}}}

\newcommand{\xins}{x \in S}

\newcommand{\pxy}{p_{xy}} 

\newcommand{\cc}[1]{\citet{#1}}

  \title{Reinforcement Learning: a Comparison of 
  UCB Versus Alternative Adaptive Policies}
\small
\author{{\bf 
Wesley Cowan} \\
   Computer Science Department, Rutgers University\\
110 Frelinghuysen Rd., Piscataway, NJ 08854 
\and {\bf Michael N. Katehakis}\\
Management Science and Information Systems Department, Rutgers University\\
 100 Rockafeller Road, Piscataway, NJ 08854, USA
\and {\bf  Daniel Pirutinsky}\\ Management Science and Information Systems Department, Rutgers University\\
 100 Rockafeller Road, Piscataway, NJ 08854, USA
}
 
\begin{document}
\maketitle
\begin{abstract}
In this paper we consider the basic version of Reinforcement Learning (RL) that involves 
computing optimal data driven (adaptive) policies for Markovian decision process with  unknown transition probabilities.  We provide a brief survey of the state of the art of the area and 
 we compare the performance of the classic UCB policy of \cc{bkmdp97} with a new policy developed herein which we call MDP-Deterministic Minimum Empirical Divergence (MDP-DMED), and a method based on Posterior sampling (MDP-PS). 
\end{abstract}


\section{Introduction} 

Reinforcement Learning (RL) refers to machine learning (ML) techniques designed   
 for sequential decision making when an agent needs to 
  ``learn'' a policy which maximizes  a  reward  (or minimizes  a cost) criterion when some parameters of the model are not known in advance, c.f.  \cc{bertsekas2019RL, sutton2018reinforcement}, \cc{mohri2018foundations,alpaydin2014introduction},  \cc{tewari2008optimistic,Tewari:Bartlett:08},  \cc{ortner2014regret}. Reinforcement learning  is experiencing significant  growth in recognition due to successful applications    in many areas c.f.    
\cc{Wiering2018}, \cc{russo2018tutorial}, \cc{chang2006adaptive}, \cc{neu2010online},     \cc{munos2016safe},  
\cc{szepesvari2009algorithms}, \cc{szepesvari2010algorithms}, \cc{filippi2010optimism}, and \cc{tewari2008optimistic,Tewari:Bartlett:08}.

In this paper we consider the basic version of a  probabilistic sequential 
decision system the  discrete time, finite state and action 
Markovian decision process (MDP) cf. \cc{dynk-y79}. 
  After a very brief survey of the state of the art of the area of 
computing optimal data driven (adaptive) policies for MDPs with  unknown transition probabilities.  Then, we compare the performance of the classic UCB policy of \cc{bkmdp97} with a new policy developed herein which we call MDP-Deterministic Minimum Empirical Divergence (MDP-DMED), and a method based on Posterior sampling (MDP-PS). The  MDP-DMED algorithm is inspired by  the DMED  method for the Multiarmed Bandit Problem developed in \cc{honda2010, honda2011} and is based on estimating the optimal rates at which actions should be taken. The  MDP-PS method is based on ideas of greedy posterior sampling that go back to \cc{Thompson}, cf. \cc{osband2017posterior}. Indeed many modern ideas of RL originate in work done for the multi-armed bandit problem cf. \cc{gittins79, gittins2011}, \cc{auer2002finite} \cc{Whi80}, \cc{Weber92}, 
\cc{villar2015multi}    \cc{sonin2011generalized, sonin2016continue},  \cc{mahajan2008multi}, \cc{Kat87, KatR96,  katehakis1986computing}. 

Some additional related work and areas of potential applications are contained in 
\cc{CowanK2019}, \cc{CowanK2015gr} \cc{azar2017minimax},  \cc{katehakis_smit_spieksma_2016},  \cc{CowanK2015gd}
\cc{abbeel2004apprenticeship}, \cc{katehakis_smit_spieksma_2016}, 
\cc{ferreira2018online}, 
\cc{jaksch2010near},  \cc{asmussen2007stochastic}.

\section{Formulation} 
A finite 
MPD  is specified by a quadruple  
$(S,A,R,P)$, where 
$S=\{1,2,\ldots ,s\}$ is the state space,
$A=\bigcup_{\xins} A(x)$ is the action space,  with  $A(x)$  being the
set of admissible actions  (or controls) in state $x$,
$R=[r_x(a)]_{\xins, a \in A(x)}$, 
is the reward structure and 
$P=[\pxy(a)]_{x,y \in S, a \in A(x)}$
is the transition law. Here    $r_x(a)$ and  $\pxy(a) $
are respectively the  one step expected reward and  transition 
probability from 
state $x$  to  state $y$ under action 
$a . $ 
For extensions regarding state and action spaces and 
continuous time we refer to [13] and references therein.
 \cc{lerma},   and  \cc{dynk-y79}. 

When all elements of $(S, A, R, P)$ are known the model is said to be an MDP
with {\sl complete information} (CI-MDP). In this case, 
optimal polices 
can be  obtained via the 
appropriate version of optimality  equations, given the prevailing  optimization criterion, state -
action - time conditions and regularity assumptions c.f.   \cc{lerma}, \cc{Spieksma94}, \cc{dynk-y79}. 

When some of the elements of $(S,A,R,P)$ are unknown 
the model is said to be an MDP with  incomplete 
or {\sl partial information} (PI-MDP).

For the body of the paper, we consider the following partial information model: the transition probability vector $\underline{p}^x(a) = [p_{x,y}(a)]_{y \in S}$ is taken to be an element of parameter space 
 $$\Theta = \vbigg\{ \underline{p} \in \mathbb{R}^{\lvert S \rvert} : \sum_{y \in S} p_y = 1 , \forall y \in S, p_{y} > 0\vbig\},$$
that is, the space of all $\lvert S \rvert$-dimensional probability vectors. The restriction that each transition probability be non-negative is simply to ensure that for any control policy, the resulting Markov chain is irreducible.
Additionally, for the body of the paper we will take the reward structure $R = [r_x(a)]_{x \in S, a \in A(x)}$ to be known, and constant. Unknown or probabilistic reward structures are to be considered in future work. 

Under this model, we define a sequence of state valued random variables $X_1, X_2, X_3, \ldots$ representing the sequence of states of the MDP (taking $X_1 = x_1$ as a given initial state), and action valued random variables $A_1, A_2, \ldots$ as the action taken by the controller, action $A_t$ being taken at time $t$ when the MDP is in state $X_t$. It is convenient to define a control policy $\pi$ as a (potentially random) history dependent sequence of actions such that $\pi(t) = \pi(X_1,A_1,\ldots,X_{t-1},A_{t-1},X_t) = A_t \in A(X_t)$. We may then define the value of a policy as the total expected reward over a given horizon of action:
\begin{equation}
V_\pi(T) = \mathbb{E}\vbiggg[ \sum_{t = 1}^T r_{X_t}(A_t) \vbigg].
\end{equation}

Let $\Pi$ be the set of all feasible MDP policies $\pi$. We are interested in policies that maximize the expected reward from the MDP, in particular policies that are capable of maximizing the expected reward irrespective of the initial uncertainty that exists about the underlying MDP dynamics (i.e., for all possible $P$ under consideration). It is convenient then to define $V(T) = \sup_{\pi \in \Pi} V_\pi(T)$. We may then define the ``regret'' as the expected loss due to ignorance of the underlying dynamics,
\begin{equation}
R_\pi(T) = V(T) - V_\pi(T).
\end{equation}
We are interested in Uniformly Fast c.f. \cc{bkmdp97} policies, that achieve $R_\pi(T) = O(\ln T)$ for all feasible transition laws $P$. In this case, despite the controller's initial lack of knowledge about the underlying dynamics, she can be assured that her expected loss due to ignorance grows not only sub-linearly over time, but slower than any power of $T$. It is shown in  \cc{bkmdp97} that any uniformly fast policy has a strict lower bound of logarithmic asymptotic growth of regret, with a bound on the order coefficient in terms of the unknown transition law $P$ and the known reward structure $R$. Policies that achieve this lower bound are Asymptotically Optimal c.f. \cc{bkmdp97}; see also \cc{CowanK2019}, \cc{CowanHK2018},  \cc{bk96}, and references therein. 

It is additionally convenient to define the following notation: with a given policy $\pi$ to be understood, we denote by $T_{x}(t)$ the number of times the MDP has been in state $x$ in the first $t$ periods; we denote by $T_{x,a}(t)$ the number of times the MDP has been in state $x$ and had action $a$ taken; we denote by $T_{x,a,y}(t)$ the number of times the MDP has transitioned from $x$ to $y$ under action $a$.

In the next subsection, we consider the case of the controller having complete information (the best possible case) and use this to motivate notation and machinery for the remainder of the paper. The body of the paper is devoted to presenting and discussing three control policies that are either provably asymptotically optimal, or at least appear to be. While no proofs are presented, the results of numerical experiments are presented demonstrating the efficacy of these policies.

\subsection{The Optimal Policy Under Known Parameters}
In this section, we consider the case of complete information, when $P$ and $R$ are known. In this case, it can be shown that there is a deterministic policy, one in which the action taken at any time depends only on the current state, that realizes the maximal long term average expected reward. Letting $\Pi_D$ be the (finite) set of all such deterministic policies: 
\begin{equation} 
\phi^*(A,P)  = \max_{\pi \in \Pi_D} \lim_T \frac{V_\pi(T)}{T} = \sup_{\pi \in \Pi} \liminf_T \frac{ V_\pi(T) }{T}.
\end{equation} 
That there is such an optimal deterministic policy is a classical result cf. [6]. 

We may characterize this optimal policy in terms of the solution for $\phi = \phi^*(A,P), \underline{v} = \underline{v}(A,P)$ of the following system of optimality equations:
\begin{equation}\label{eqn:dp}
  \forall x \in S:\ 
 \ \ \ \ \   \phi + v_x = \max_{a \in A(x)}\vbiggg\{  r_x(a) + \sum_{y \in S} p_{x,y}(a) v_y\vbiggg\}.
\end{equation}

Given the solution $\phi$ and vector $\underline{v}$ to the above equations, the asymptotically optimal policy $\pi^*$ can be characterized as, whenever in state $x \in S$, take any action $a$ for which 
\begin{equation}
\phi + v_x = r_x(a) + \sum_{y \in S} p_{x,y}(a) v_y.
\end{equation}
We denote the set of such asymptotically optimal actions as $O(x,P)$. In general, $a^*(x,P)$ should be taken to denote an action $a^* \in O(x,P)$.

The solution $\phi$ above represents the maximal long term average expected reward. The vector $\underline{v}$, i.e., $v_x$ for any $x \in S$, represents in some sense the immediate value of being in state $x$ \textit{relative to the long term average expected reward}. The value $v_x$ essentially encapsulates the future opportunities for value available due to being in state $x$. 

It will be convenient in what is to follow to define the following notation:
\begin{equation}
L(x,a,\underline{p}, \underline{v}) = r_x(a) + \sum_{y \in S} p_y v_y.
\end{equation}
The function $L$ effectively represents the value of a given action in a given state, for a given transition vector - both the immediate reward, and the expected future value of whatever state the MDP transitions into. The value of an asymptotically optimal action for any state $x$ is thus given by $L^*(x,A,P) = L(x, a^*(x,P),\underline{p}^x(a^*(x,P)), \underline{v}(A,P))$. It can be  shown that the ``expected loss'' due to an asymptotically sub-optimal action, taking action $a \notin O(x,P)$ when the MDP is in state $x$, is effectively in the limit given by
\begin{equation}
 \Delta(x,a,A,P)  = L^*(x,A,P) - L(x,a, \underline{p}^x(a), \underline{v}(A,P)).
\end{equation}

In the general (partial or complete information) case, it is shown in  [6] that the regret of a given policy $\pi \in \Pi$ can be expressed asymptotically as
\begin{equation}
 R_\pi(T)   = \sum_{x \in S} \sum_{a \notin O(x,P)} \mathbb{E}\left[ T_{x,a}(T) \right] \Delta(x,a,A,P) + O(1).
\end{equation}

Note, the above formula justifies the description of $\Delta(x,a,A,P)$ as the ``average loss due to sub-optimal activation of $a$ in state $x$'. Additionally, from the above it is clear that in the case of complete information, when $P$ is known and therefore the asymptotically optimal actions are computable, the total regret at any time $T$ is bound by a constant. Any expected loss at time $T$ is due only to finite horizon effects. In general, for the incomplete information case, we have the following bound due to [6], for any uniformly fast policy $\pi$,
\begin{equation}
\liminf_T \frac{ R_\pi(T) }{\ln T} \geq \sum_{x \in S} \sum_{a \notin O(x,P)} \frac{ \Delta(x,a,A,P) }{ \mathbf{K}_{x,a}(P) },
\end{equation}
where $\mathbf{K}_{x,a}(P)$ represents the minimal Kullback-Leibler divergence between $\underline{p}^x(a)$ and any $\underline{q} \in \Theta$ such that substituting $\underline{q}$ for $\underline{p}^x(a)$ in $P$ renders $a$ the unique optimal action for $x$. Note, the Kullback-Leibler divergence is given by $\mathbf{I}(\underline{p}, \underline{q}) = \sum_{x \in S} p_x \ln( p_x / q_x )$. Policies that achieve this lower bound, for all $P$, are referred to as Asymptotically Optimal. 


\section{The UCB Algorithm for MDPs Under Unknown  Transition Distributions} \label{sec:UCB}
The policy we present here is a simplified version of the UCB-MDP policy developed  in \cc{bkmdp97}.
In this classical upper confidence MDP-UCB setting   in each time instance 
estimates of the values of each available action are computed based on available data, inflated by a certain confidence interval (based on the Kullback-Leibler divergence). The more data on a given action that is available, the tighter the confidence interval will be, and therefore the less the corresponding estimate will be inflated.

At any time $t \geq 1$, let $x_t$ be the current (given) state of the MDP. We construct the following estimators:
\begin{itemize}
\item Transition Probability Estimators: for each state $y$ and action $a \in A(x_t)$, construct $\hat{P}_t$ based on
\begin{equation}
\hat{p}_{x_t,y}(a) = \frac{ T_{x_t,a,y}(t) + 1 }{ T_{x_t,a}(t) + \lvert S \rvert }.
\end{equation}
Note, the biasing terms (the $1$ in the numerator, $\lvert S \rvert$ in the denominator) serve to force the estimated transition probabilities away from $0$, and thus our estimates of $\underline{p}^{x_t}(a)$ will be in $\Theta$. 

\item ``Good'' Action Sets: construct the following subset of the available actions $A(x_t)$,
\begin{equation}
\hat{A}_t = \vbigg\{ a \in A(x_t) : T_{x_t,a}(t) \geq \left( \ln T_{x_t}(t) \right)^2 \vbigg\}.
\end{equation}
The set $\hat{A}_t$ represents the actions available from state $x_t$ that have been sampled frequently enough that the estimates of the associated transition probabilities should be ``good'. In the limit, we expect that sub-optimal actions will be taken only logarithmically, and hence for sufficiently large $t$, $\hat{A}_t$ will contain only actions that are truly optimal. If no actions have been taken sufficiently many times, we take $\hat{A}_t = A(x_t)$ to prevent it from being empty.

\item Value Estimates: having constructed these estimators, we compute $\hat{\phi}_t = \phi(\hat{A}_t, \hat{P}_t)$ and $\hat{\underline{v}}_t = \underline{v}(\hat{A}_t, \hat{P}_t)$ as the solution to the optimality equations in Eq. \eqref{eqn:dp}, essentially treating the estimated probabilities as correct and computing the optimal values and policy for the resulting estimated MDP.
\end{itemize}

At this point, we implement the following UCB index based decision rule: for each action $a \in A(x_t)$, we compute the following \textit{index}:
\begin{equation} \label{eq:UCB-index}
u_a(t) = \sup_{\underline{q} \in \Theta} \vbiggg\{ L(x_t, a, \underline{q}, \hat{\underline{v}}) : \mathbf{I}(\underline{\hat{p}}^{x_t}, \underline{q}) \leq \frac{ \ln t}{T_{x_t,a}(t) } \vbiggg\},
\end{equation}
where $\mathbf{I}(\underline{p}, \underline{q}) = \sum_y p_y \ln( p_y / q_y )$ is the Kullback-Leibler divergence, and take action 
\begin{equation}
\pi(t) = \argmax_{a \in A(x_t)}  \vbig\{u_a(t)  \vbigg\}.
\end{equation}

This is a natural extension of several classical KL-divergence based UCB policies for the multi-armed bandit problem cf. \cc{CowanK2019}, \cc{bk96}, and references therein,  taking the view of the $L$ function as the ``value'' of taking a given action in a given state, estimated with the current data. In \cc{bkmdp97}, a   version of the above policy is in fact shown to be asymptotically optimal. The modification is largely for analytical benefit however, the pure UCB index policy defined above shows excellent performance  cf. Figure 1. Further discussion of the performance of this policy is given in the Comparison of Performance section.

An important and legitimate concern to the practical usage of this UCB policy that has been noted in \cc{tewari2008optimistic} among others, is actually calculating the index in Eq. \eqref{eq:UCB-index}. Efficient formulations can be derived and this will be explored in depth in future work.

\section{A DMED-Type Algorithm for MDPs Under Uncertain Transitions}\label{sec:DMED}
In the classical DMED algorithm for Multi-armed Bandit Problems, the decision process proceeds by attempting to successively estimate the asymptotically minimal rates with which sub-optimal actions must be taken, and then attempting to take actions in such a way so as to realize the estimated minimal rates. As applied to MDPs, we have the following relationship from [6]. For any uniformly fast policy $\pi$, for any state $x$ and sub-optimal action $a \notin O(x,P)$,
\begin{equation}
\liminf_T \frac{ \mathbb{E}\left[ T_{x,a}(T) \right] }{ \ln T} \geq \frac{1}{ \mathbf{K}_{x,a}(P) },
\end{equation}
where $\mathbf{K}_{x,a}(P)$ is, as before, the minimal Kullback-Leibler divergence $\mathbf{I}(\underline{p}^x(a),\underline{q})$ between the true transition probability vector $\underline{p}^x(a)$, and any transition probability vector $\underline{q} \in \Theta$ such that substituting $\underline{q}$ for $\underline{p}^x(a)$ in $P$ would render action $a$ uniquely optimal for state $x$.

Computing the function $\mathbf{K}_{x,a}(P)$ is not easy. We consider the following substitute, then:
\begin{equation}
 \mathbf{\tilde{K}}_{x,a}(P, \underline{v}, a^*) =  
  \inf_{\underline{q} \in \Theta} \vbigg\{ \mathbf{I}(\underline{p}^x(a), \underline{q}): L(x,a,\underline{q}, \underline{v}) > L(x,a^*,\underline{p}^x(a^*), \underline{v}) \vbigg\}.
\end{equation}

The function $\mathbf{K}$ measures how far the transition vector associated with $x$ and $a$ must be perturbed (under the KL-divergence) to make $a$ the optimal action for $x$. The function $\mathbf{\tilde{K}}$ measures how far the transition vector associated with $x$ and $a$ must be perturbed (under the KL-divergence) to make the value of $a$, as measured by the $L$-function, greater than the value of an optimal action $a^*$.

In this way, we have the following approximate MDP-DMED algorithm; see \cc{honda2010, honda2011}
for a multi-armed bandit version of this policy.

At any time $t \geq 1$, let $x_t$ be the current state, and construct the estimators as in the UCB-MDP algorithm in section \ref{sec:UCB}, $\hat{P}_t$, $\hat{A}_t$, and utilize these to compute the estimated optimal values, $\hat{\phi}_t = \phi(\hat{A}_t, \hat{P}_t)$ and $\underline{\hat{v}}_t = \underline{v}( \hat{A}_t, \hat{P}_t)$.

Let $\hat{a}^*_t = \argmax_{a \in A(x_t)} L(x_t, a, \underline{\hat{p}}^{x_t}(a), \underline{v}_t)$ be the estimated ``best'' action to take at time $t$. For each $a \neq \hat{a}^*_t$, compute the discrepancies $$D_t(a) = \ln t / \mathbf{\tilde{K}}_{x_t,a}(\hat{P}_t, \underline{\hat{v}}_t, \hat{a}^*_t) - T_{x_t,a}(t).$$

If $\max_{a \neq \hat{a}^*_t} D_t(a) \leq 0$, take $\pi(t) = \hat{a}^*_t$, otherwise, take $\pi(t) = \argmax_{a \neq \hat{a}^*_t} D_t(a).$

Following this algorithm, we perpetually reduce the discrepancy between the estimated sub-optimal actions, and the estimated rate at which those actions should be taken. The exchange from $\mathbf{K}$ to $\mathbf{\tilde{K}}$ sacrifices some performance in the pursuit of computational simplicity, however it also seems clear from computational experiments that DMED-MDP as above is not only computationally tractable, but also produces reasonable performance in terms of achieving small regret  cf.  Figure 1. Further discussion of the performance of this policy is given in the Comparison of Performances section.

\section{A Posterior Sampling Algorithm for MDPs}
In this section we introduce a  Posterior Sampling (Thompson-Type ) policy for MDPs, or PS-MDP. This type of policy is also known as Thompson Sampling, or Probability matching. The basic idea is to generate estimates for the unknown parameters (transition probabilities) randomly, according to the posterior distribution for those unknown parameters, based on the current data. 
In particular, PS-MDP proceeds in the following way:

At any time $t \geq 1$, let $x_t$ be the current state of the MDP. As in UCB-MDP and DMED-MDP previously, construct the estimators $\hat{P}_t, \hat{A}_t, \hat{\phi}_t, \underline{v}_t$. In addition, generate the following random vectors.

For each action $a \in A(x_t)$, let $\underline{T}_{x_t,a}(t) = [ T_{x_t,a,y}(t) ]_{y \in S}$ be the vector of observed transition counts from state $x_t$ to $y$ under action $a$. Generate the random vector $\underline{Q}$ according to
\begin{equation}
\underline{Q}^{a}(t) \sim \text{Dir}( \underline{T}_{x_t,a}(t) ).
\end{equation}
The $\underline{Q}^a(t)$ are distributed according to the joint posterior distribution of $\underline{p}^{x_t}(a)$ with a uniform prior.

At this point, define the following values as posterior estimates of the potential value of each action:
\begin{equation}
W_a(t) = r_{x_t}(a) + \sum_{y} Q^a_y(t) \hat{v}_y,
\end{equation}
and take action $\pi(t) = \argmax_{a \in A(x_t)} W_a(t).$

%
%
%
%

\section{Comparison of Performance} \label{sec:comparison}
In this section we discuss the results of our simulation test of these policies on a small example  with 3 states ($x_1,x_2,$ and $x_3$) with 2 available actions ($a_1$ and $a_2$) in each state. Below we show the transition probabilities, as well as the reward, returned under each action.
 
{\small
\begin{center}

$P[a_1] = \ \ \  $ \begin{tabular}{|l | c | c | c | }
	& $x_1$ & $x_2$ & $x_3$ \\ \hline
	$x_1$ & 0.04 & 0.69 & 0.27 \\ \hline
	$x_2$ & 0.88 & 0.01 & 0.11 \\ \hline
	$x_3$ & 0.02 & 0.46 & 0.52 \\ 
\end{tabular}
\end{center}
\begin{center}
$P[a_2] =  \ \ \  $  \begin{tabular}{|l | c | c | c | }
	& $x_1$ & $x_2$ & $x_3$ \\ \hline
	$x_1$ & 0.28 & 0.68 & 0.04 \\ \hline
	$x_2$ & 0.26 & 0.33 & 0.41 \\ \hline
	$x_3$ & 0.43 & 0.35 & 0.22 \\ 
\end{tabular}
\end{center}

\begin{center}
\ \ \	$R=  \ \ \  $  \begin{tabular}{|l | c | c | c | }
		& $x_1$ & $x_2$ & $x_3$ \\ \hline
		$a_1$ & 0.13 & 0.47 & 0.89   \\ \hline
		$a_2$ & 0.18 & 0.71 & 0.63 \\
	\end{tabular} 
\end{center}
}
If these transition probabilities were known, the optimal policy for this MDP would be  $\pi^*(x_1) = a_1, \pi^*(x_2) = a_2,$ and $ \pi^*(x_3) =  a_1$.

\begin{figure}\label{fig:cumulative-regret}
	\begin{center}
		\includegraphics[scale=.15]{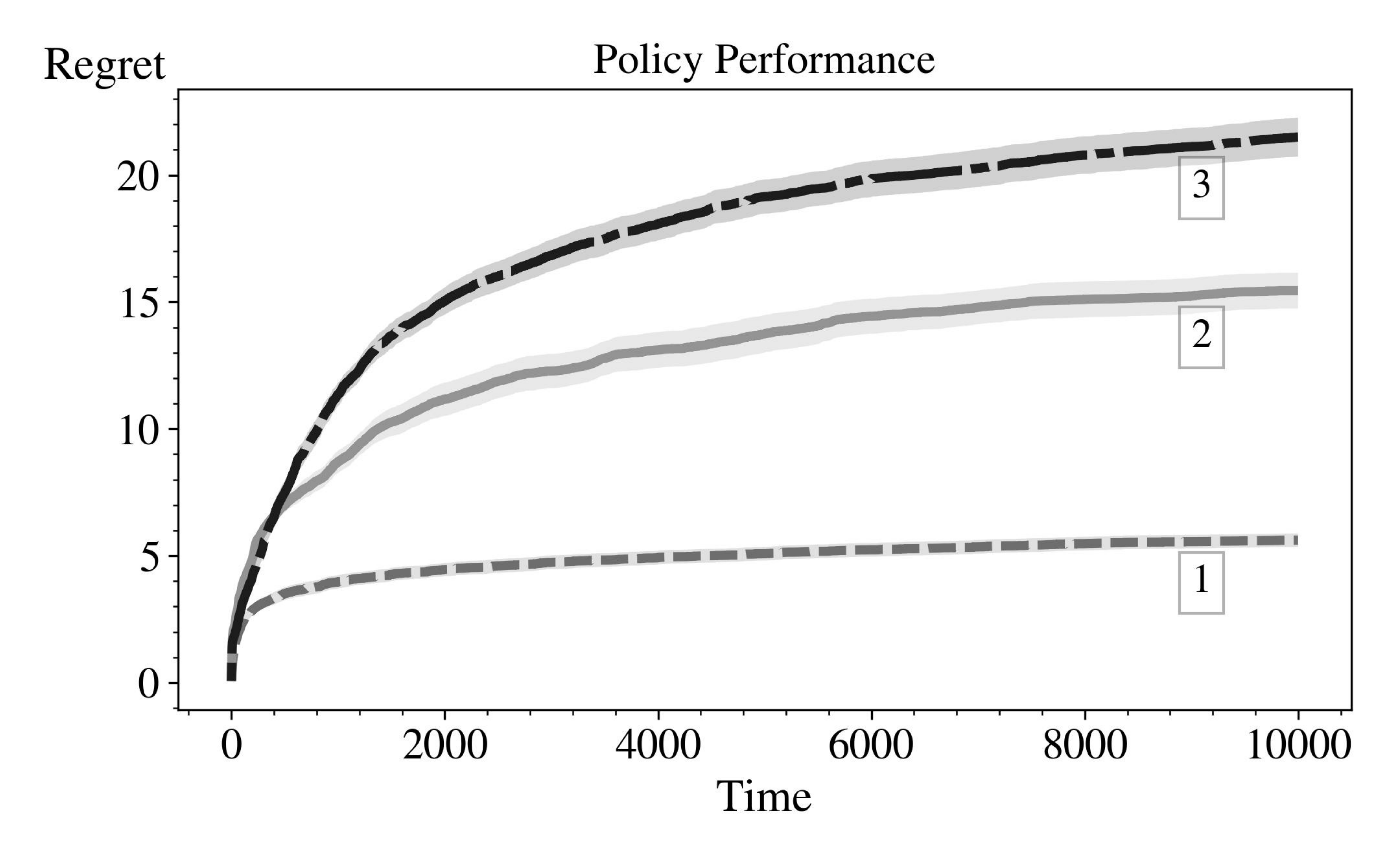}	
		\caption{\small Average cumulative regret over time for each policy \label{fig:simulation}} 
	\end{center}
\end{figure}

We simulated each policy 100 times over a time horizon of 10,000 and for each time step we computed the mean regret as well as the variance. In Figure \ref{fig:simulation}, we plot the mean regret over time for each policy, [1] PS, [2] UCB, and [3] DMED, along with a $95\%$ confidence interval for all sample paths.

We can see that all policies seem to have logarithmic growth of regret. There are a few interesting differences that the plot highlights, at least for these specific parameter values:

DMED-MDP has not only the highest finite time regret, but also large variance that seems to increase over time. This seems primarily due to the ``epoch'' based nature of the policy, which results in exponentially long periods when the policy may get trapped taking sub-optimal actions, incurring large regret until the true optimal actions are discovered. The benefit of this epoch structure is that once the optimal actions are discovered, they are taken for exponentially long periods, to the exclusion of sub-optimal actions.

PS-MDP seems to perform best, exhibiting lowest finite time regret as well as the tightest variance. This seems largely in agreement with the performance of PS-type policies in other bandit problems as well, in which they are frequently asymptotically optimal cf. \cc{Agrawal2013ts, Agrawal2012ts}, \cc{honda13ts}, \cc{Kaufmann2012} and references therein. 


\subsection{Policy Robustness - Inaccurate Priors}

How do these policies respond to potentially ``unlucky'' or non-representative streaks of data? Can these policies be fooled, and what are the resulting costs before they recover? 

To test the robustness of  these policies, with respect to prior information, we ``rigged'' the first 60 actions and transitions, such that under the estimated transition probabilities the optimal policy would be to activate the sub-optimal action in each state. In more detail, let $T_{x,y}(a)$ be the number of times we transitioned from state $x$ to state $y$ under action $a$. Then we ``rigged''  $T(a)$ so that it started like so,

\begin{center}
	
	$T[a_1] =\ \ \ $ \begin{tabular}{| l | c | c | c | }
		& $x_1$ & $x_2$ & $x_3$ \\ \hline
		$x_1$ & 8 & 1 & 1 \\ \hline
		$x_2$ & 1 & 1 & 8 \\ \hline
		$x_3$ & 8 & 1 & 1 \\ 
	\end{tabular} \ ,\\
\end{center}
\begin{center}
	$T[a_2] =\ \ \ $ \begin{tabular}{| l | c | c | c | }
		& $x_1$ & $x_2$ & $x_3$ \\ \hline
		$x_1$ & 1 & 1 & 8 \\ \hline
		$x_2$ & 8 & 1 & 1 \\ \hline
		$x_3$ & 1 & 1 & 8 \\ 
	\end{tabular}
\end{center}

%
%
Under the resulting (bad) estimated transition probabilities, we have that the optimal policy is $\hat{\pi}^*(x_1) = a_2,  \hat{\pi}^*(x_2) = a_1$, and $\hat{\pi}^*(x_3) = a_1$. 
Under these initial estimates, the assumed optimal policy chooses the sub-optimal action in each state.

\begin{figure}\label{fig:stratified-regret}
	\begin{center}
		\includegraphics[scale=.15]{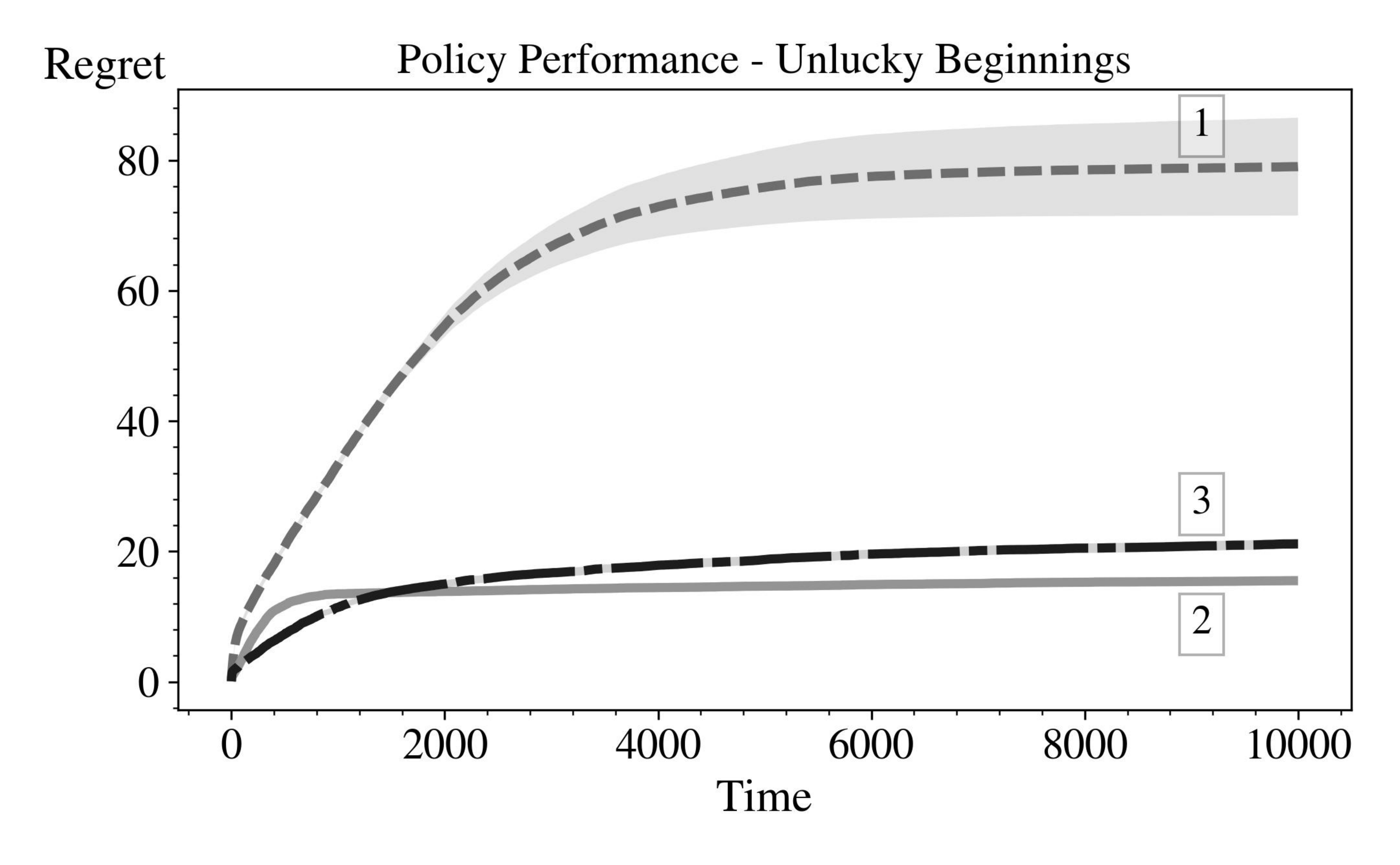}	
		\caption{\small Robustness test. UCB seems to be largely unaffected by the unlucky beginning.  \label{fig:robustness}} 
	\end{center}
\end{figure}

The subsequent performances of the MDP policies are plotted in Figure \ref{fig:robustness}. All policies still appear  to have logarithmic growth in regret, suggesting they can all ``recover'' from the initial bad estimates. It is striking though, the extent to which the average regrets for DMED-MDP and PS-MDP are affected, increasing dramatically as a result, PS-MDP demonstrating an increase in variance as well. However, the UCB-MDP policy seems relatively stable: its average regret has barely increased, and maintains a small variance. Empirically, this phenomenon appears common for the UCB-MDP policy under other extreme conditions.

\section*{Acknowledgments}
We acknowledge support for this work from the National Science Foundation, NSF grant CMMI-1662629.

\bibliographystyle{natbib}
\vspace{-.6cm}
\bibliographystyle{plainnat}


\begin{thebibliography}{}

\bibitem[Abbeel and Ng(2004)]{abbeel2004apprenticeship}
Abbeel, P. and Ng, A.~Y. (2004).
\newblock Apprenticeship learning via inverse reinforcement learning.
\newblock In {\em Proceedings of the twenty-first international conference on
  Machine learning}, page~1. ACM.

\bibitem[Agrawal S and Goyal N.(2012)]{Agrawal2012ts}
 Agrawal, S. and  Goyal, N. (2012).
 \newblock Analysis of Thompson sampling for the multi-armed bandit problem. 
\newblock   {\em  In Conference on Learning Theory}   39--1, Springer.


\bibitem[Agrawal S and Goyal N.(2012)]{Agrawal2013ts}
 Agrawal, S. and  Goyal, N. (2013).
 \newblock  Further optimal regret bounds for Thompson sampling.
\newblock   {\em  In Artificial intelligence and statistics}   99--107.

\bibitem[Alpaydin(2014)]{alpaydin2014introduction}
Alpaydin, E. (2014).
\newblock {\em Introduction to machine learning}.
\newblock MIT press.

\bibitem[Asmussen and Glynn(2007)]{asmussen2007stochastic}
Asmussen, S. and Glynn, P.~W. (2007).
\newblock {\em Stochastic simulation: algorithms and analysis}, volume~57.
\newblock Springer Science \& Business Media.

\bibitem[Auer {\em et~al.}(2002)]{auer2002finite}
Auer, P., Cesa-Bianchi, N., and Fischer, P. (2002).
\newblock Finite-time analysis of the multiarmed bandit problem.
\newblock {\em Machine learning}, {\bf 47}(2-3), 235--256.

\bibitem[Azar {\em et~al.}(2017)]{azar2017minimax}
Azar, M.~G., Osband, I., and Munos, R. (2017).
\newblock Minimax regret bounds for reinforcement learning.
\newblock {\em arXiv preprint arXiv:1703.05449}.

\bibitem[Bertsekas (2019)]{bertsekas2019RL}
Bertsekas, D.~P. (2019).
\newblock Reinforcement learning and optimal control.
\newblock {\em Athena Scientific, Belmont, Massachusetts}.

\bibitem[Burnetas and Katehakis(1997)]{bkmdp97}
Burnetas, A.~N. and Katehakis, M.~N. (1997).
\newblock Optimal adaptive policies for {M}arkov decision processes.
\newblock {\em Mathematics of {O}perations {R}esearch}, {\bf 22}(1), 
222--255.
  \bibitem[Burnetas and Katehakis(1996)]{bk96}
Burnetas, A.~N. and Katehakis, M.~N. (1996).
\newblock 
  Optimal Adaptive Policies for Sequential Allocation Problems.
  \newblock {\em Advances in Applied Mathematics}, 17 (2)   122--142.

\bibitem[Chang {\em et~al.}(2006)]{chang2006adaptive}
Chang, M., Chow, S.-C., and Pong, A. (2006).
\newblock Adaptive design in clinical research: issues, opportunities, and
  recommendations.
\newblock {\em Journal of biopharmaceutical statistics}, {\bf 16}(3), 299--309.

\bibitem[Cowan and Katehakis(2019)]{CowanK2019}
Cowan W., and M.N. Katehakis (2019).  
\newblock Exploration–exploitation policies with almost sure, arbitrarily slow growing asymptotic regret, DOI={10.1017/S0269964818000529}.
\newblock {\em Probability in the Engineering and Informational Sciences},   1--23.

\bibitem[Cowan {\em et~al.}(2018)]{CowanHK2018}
Cowan W., Honda Y. and M.N. Katehakis (2018). 
\newblock Normal Bandits of Unknown Means and Variances: 
Asymptotic Optimality, Finite Horizon Regret Bounds, and a Solution to an Open Problem, \newblock {\em Journal of Machine Learning Research}(JMLR), 18, 1--28.

\bibitem[Cowan and Katehakis(2015)]{CowanK2015gr}
Cowan W., and M.N. Katehakis (2015).  
\newblock   Asymptotically Optimal Sequential Experimentation Under Generalized Ranking.
\newblock  arXiv:1510.02041  

\bibitem[Cowan and Katehakis(2015)]{CowanK2015gd}
Cowan W., and M.N. Katehakis (2015).  
\newblock   Multi-armed Bandits under General Depreciation and Commitment, 
\newblock {\em Probability in the Engineering and Informational Sciences}, 29 (1), 51--76. 

\bibitem[Dekker {\em et~al.}(1994)]{Spieksma94}
Dekker, R., Hordijk, A., and Spieksma, F.~M. (1994).
\newblock On the relation between recurrence and ergodicity properties in
  denumerable {M}arkov decision chains.
\newblock {\em Mathematics of Operations Research}, {\bf 19}, 3.

\bibitem[Dynkin and Yushkevich(1979)]{dynk-y79}
Dynkin, E. and Yushkevich, A. (1979).
\newblock {\em Controlled {M}arkov processes}, volume 235.
\newblock Springer.

\bibitem[Ferreira {\em et~al.}(2018)]{ferreira2018online}
Ferreira, K.~J., Simchi-Levi, D., and Wang, H. (2018).
\newblock Online network revenue management using thompson sampling.
\newblock {\em Operations research}, {\bf 66}(6), 1586--1602.

\bibitem[Filippi {\em et~al.}(2010)]{filippi2010optimism}
Filippi, S., Capp{\'e}, O., and Garivier, A. (2010).
\newblock Optimism in reinforcement learning based on {K}ullback {L}eibler
  divergence.
\newblock In {\em 48th Annual Allerton Conference on Communication, Control,
  and Computing}.

\bibitem[Henderson {\em et~al.}(2018)]{henderson2018deep}
Henderson, P., Islam, R., Bachman, P., Pineau, J., Precup, D., and Meger, D.
  (2018).
\newblock Deep reinforcement learning that matters.
\newblock In {\em Thirty-Second AAAI Conference on Artificial Intelligence}.

\bibitem[Lerma(2012)]{lerma}
Lerma, H.O. (2012).
\newblock {\em Adaptive Markov control processes}, volume~79.
\newblock Springer Science \& Business Media.

\bibitem[Gittins(1979)]{gittins79}
Gittins, J. (1979)
\newblock Bandit processes and dynamic allocation indices (with discussion).
\newblock \emph{J. Roy. Stat. Soc. Ser. B}, 41:\penalty0 335--340.

\bibitem[Gittins et~al.(2011)Gittins, Glazebrook, and Weber]{gittins2011}
John~C. Gittins, Kevin Glazebrook, and Richard~R. Weber.
\newblock \emph{Multi-armed Bandit Allocation Indices}.
\newblock John Wiley \& Sons, West Sussex, U.K..

\bibitem[Honda and Takemura(2010)]{honda2010}
  Honda J. and   Takemura A. (2010). 
\newblock An asymptotically optimal bandit algorithm for bounded support
  models.
\newblock In \emph{COLT},   67--79.

\bibitem[Honda and Takemura(2011)]{honda2011}
  Honda J. and   Takemura A. (2011). 
\newblock An asymptotically optimal policy for finite support models in the
  multiarmed bandit problem.
\newblock \emph{Machine Learning}, 85\penalty0 (3):\penalty0 361--391.

\bibitem[Honda and Takemura(2013)]{honda13ts}
  Honda J. and   Takemura A. (2013). 
\newblock Optimality of {T}hompson sampling for {G}aussian bandits depends on
  priors.
\newblock arXiv preprint arXiv:1311.1894.

\bibitem[Jaksch {\em et~al.}(2010)]{jaksch2010near}
Jaksch, T., Ortner, R., and Auer, P. (2010).
\newblock Near-optimal regret bounds for reinforcement learning.
\newblock {\em Journal of Machine Learning Research}, {\bf 11}(Apr),
  1563--1600.
  
  
\bibitem[{Katehakis and Derman(1986)}]{katehakis1986computing}
Katehakis, Michael~N. and C.~Derman (1986). \newblock{Computing optimal
  sequential allocation rules in clinical trials.} \emph{Lecture
  Notes-Monograph Series}, 29 -- 39.
\bibitem[{Katehakis and Rothblum(1996)}]{KatR96}
Katehakis, Michael~N. and Uriel~G. Rothblum (1996). \newblock {Finite state
  multi-armed bandit problems: Sensitive-discount, average-reward and
  average-overtaking optimality.} \emph{The Annals of Applied Probability}, 6,
  1024--1034.
 

\bibitem[{Katehakis and Veinott~Jr(1987)}]{Kat87}
Katehakis, Michael~N. and Arthur~F Veinott~Jr (1987). \newblock{The multi-armed
  bandit problem: decomposition and computation.} \emph{Math. Oper. Res.}, 12,
  262 -- 68.
 
\bibitem[Katehakis {\em et~al.}(2016)]{katehakis_smit_spieksma_2016}
Katehakis, M., Smit, L.~C., and Spieksma, F. (2016).
\newblock A comparative analysis of the successive lumping and the lattice path
  counting algorithms.
\newblock {\em Journal of Applied Probability}, {\bf 53}(1), 106--120.

\bibitem[Kaufmann {\em et~al.}(2016)]{Kaufmann2012}
Kaufmann, E., Korda, N. and Munos, R., (2012).
\newblock   Thompson sampling: An asymptotically optimal finite-time analysis. 
\newblock {\em In International Conference on Algorithmic Learning Theory},   199--213). Springer, Berlin, Heidelberg.


\bibitem[Mahajan and Teneketzis(2008)]{mahajan2008multi}
Mahajan, A. and Teneketzis, D. (2008).
\newblock Multi-armed bandit problems.
\newblock In {\em Foundations and Applications of Sensor Management},  
 121--151. Springer.

\bibitem[Mohri {\em et~al.}(2018)]{mohri2018foundations}
Mohri, M., Rostamizadeh, A., and Talwalkar, A. (2018).
\newblock {\em Foundations of machine learning}.
\newblock MIT press.

\bibitem[Munos {\em et~al.}(2016)]{munos2016safe}
Munos, R., Stepleton, T., Harutyunyan, A., and Bellemare, M. (2016).
\newblock Safe and efficient off-policy reinforcement learning.
\newblock In {\em Advances in Neural Information Processing Systems},  
  1054--1062.

\bibitem[Neu {\em et~al.}(2010)]{neu2010online}
Neu, G., Antos, A., Gy{\"o}rgy, A., and Szepesv{\'a}ri, C. (2010).
\newblock Online markov decision processes under bandit feedback.
\newblock In {\em Advances in Neural Information Processing Systems},  
  1804--1812.

\bibitem[Ortner {\em et~al.}(2014)]{ortner2014regret}
Ortner, R., Ryabko, D., Auer, P., and Munos, R. (2014).
\newblock Regret bounds for restless Markov bandits.
\newblock {\em Theoretical Computer Science}, {\bf 558}, 62--76.

\bibitem[Osband and Van~Roy(2017)]{osband2017posterior}
Osband, I. and Van~Roy, B. (2017).
\newblock Why is posterior sampling better than optimism for reinforcement
  learning?
\newblock In {\em Proceedings of the 34th International Conference on Machine
  Learning-Volume 70},   2701--2710. JMLR. org.

\bibitem[Russo and Van~Roy(2014)]{russo2018tutorial}
Russo, D.~J. and Van~Roy, B. (2014).
\newblock Learning to optimize via posterior sampling.
\newblock {\em Mathematics of Operations Research}, {\bf 39}(4), 1221–1243.

\bibitem[{Sonin(2008)}]{sonin2008generalized}
Sonin, I.M. (2008). \newblock{A generalized {G}ittins index for a {M}arkov chain
  and its recursive calculation.} \emph{Statistics \& Probability Letters}, 78,
  1526 -- 1533.
 

\bibitem[{Sonin(2011)}]{sonin2011generalized}
Sonin, I.M. (2011). \newblock{Optimal stopping of {M}arkov chains and three
  abstract optimization problems.} \emph{Stochastics An International Journal
  of Probability and Stochastic Processes}, 83, 405 -- 414.
 

\bibitem[{Sonin and Steinberg(2016)}]{sonin2016continue}
Sonin, Isaac~M and Constantine Steinberg (2016). \newblock{Continue, quit,
  restart probability model.} \emph{Annals of Operations Research}, 241,
  295--318.
  
\bibitem[Sutton and Barto(2018)]{sutton2018reinforcement}
Sutton, R. and Barto, A. (2018).
\newblock {\em Reinforcement learning: An introduction}.
\newblock MIT press.

\bibitem[Szepesv{\'a}ri(2009)]{szepesvari2009algorithms}
Szepesv{\'a}ri, C. (2009).
\newblock Algorithms for reinforcement learning.
\newblock {\em Morgan and Claypool}.

\bibitem[Szepesv{\'a}ri(2010)]{szepesvari2010algorithms}
Szepesv{\'a}ri, C. (2010).
\newblock Algorithms for reinforcement learning.
\newblock {\em Synthesis Lectures on Artificial Intelligence and Machine
  Learning}, {\bf 4}(1), 1--103.

\bibitem[Tewari and Bartlett(2008a)]{tewari2008optimistic}
Tewari, A. and Bartlett, P. (2008a).
\newblock Optimistic linear programming gives logarithmic regret for
  irreducible {MDPs}.
\newblock In {\em Advances in Neural Information Processing Systems},  
  1505 -- 1512.

\bibitem[Tewari and Bartlett(2008b)]{Tewari:Bartlett:08}
Tewari, A. and Bartlett, P. (2008b).
\newblock Optimistic linear programming gives logarithmic regret for
  irreducible {MDPs}.
\newblock In Y.~S. J.C.~Platt, D.~Koller and S.~Roweis, editors, {\em Advances
  in Neural Information Processing Systems}, volume~20,   1505 -- 1512.
  NIPS, New York.
\bibitem[Thompson(1933]{Thompson}
Thompson W.R. (1933).  
\newblock  On the likelihood that one unknown probability exceeds another in view of the evidence of two samples. \emph{Biometrika},  25, 285--94.

\bibitem[{Villar et~al.(2015)Villar, Bowden, and Wason}]{villar2015multi}
Villar, Sof{\'\i}a~S, Jack Bowden, and James Wason (2015). \newblock{Multi-armed
  bandit models for the optimal design of clinical trials: benefits and
  challenges.} \emph{Statistical Science}, 30, 199--215.
 

\bibitem[{Weber(1992)}]{Weber92}
Weber, R. (1992). \newblock{On the {G}ittins index for multiarmed bandits.}
  \emph{The Annals of Applied Probability}, 2, 1024 -- 1033.
 

\bibitem[{Whittle(1980)}]{Whi80}
Whittle, P. (1980). \newblock{Multi-armed bandits and the {G}ittins index.}
  \emph{J. R. Statist. Soc. B}, 42, 143 -- 49.
 

\bibitem[Wiering(2018)]{Wiering2018}
Wiering, M. (2018).
\newblock Reinforcement learning: from methods to applications.
\newblock {\em Nieuw Archief voor Wiskunde}, {\bf 5}(19), 157--167.

\end{thebibliography}
\end{document}